\documentclass[lettersize,journal]{IEEEtran}
\usepackage{amsmath,amsfonts}
\usepackage{multirow}
\usepackage{booktabs}
\usepackage{algorithmic}
\usepackage{algorithm}
\usepackage[dvipsnames]{xcolor}
\usepackage{array}
\usepackage[caption=false,font=normalsize,labelfont=sf,textfont=sf]{subfig}
\usepackage{textcomp}
\usepackage{stfloats}
\usepackage{url}
\usepackage{verbatim}
\usepackage{graphicx}
\usepackage{cite}
\usepackage[percent]{overpic}
\usepackage[normalem]{ulem}
\usepackage[nolist]{acronym}

\newacro{kl}[KL]{Kullback-Leibler}
\newacro{em}[EM]{Expectation-Maximization}
\newacro{gmm}[GMM]{Gaussian Mixture Model}
\newacro{pdf}[PDF]{Probability Density Function}
\newacro{bo}[BO]{Bayesian Optimization}
\newacro{dp}[DP]{Diffusion Policy}
\newacro{bdp}[BDP]{Bayesian Diffusion Policy}

\begin{document}

\title{PAC-DP: PAC-Bayesian Diffusion Policy Learning}

\author{Mohammad Hasan Yeganegi$^{1,2}$, Dian Yu$^{2}$, Andrea Del Prete$^{1}$, Majid Khadiv$^{2}$, Matteo Saveriano$^{1}$
\thanks{$^{1}$ Department of Industrial Engineering, University of Trento, Italy. {\tt\small firstname.lastname@unitn.it}}
\thanks{$^{2}$ Munich Institute of Robotics and Machine Intelligence
(MIRMI), Technical University of Munich (TUM), Germany. {\tt\small firstname.lastname@tum.de}}}




\maketitle

\begin{abstract}

Diffusion Policies (DPs) are able to perform complex manipulation tasks. However, DPs are typically trained by minimizing a denoising objective, which provides limited control over generalization in the finite-data regimes common in robotics. In this letter, we propose \textit{PAC-DP}, an approach that increases the performance of DPs in robotic manipulation tasks. 
By modeling the DP as a Bayesian neural network, and defining a PAC-Bayes generalization bound, we derive a novel training objective that augments the standard denoising loss with a Kullback--Leibler divergence regularizer between the posterior and prior parameter distributions. From the theoretical perspective, our approach provides a principled approach to regularize the training of DPs without significantly increasing the training time. 
From the practical point of view, experimental results demonstrate improved denoising performance, lower variational negative log-likelihood, and higher success rates across multiple robotic manipulation benchmarks. Crucially, the largest improvements are observed in low-data training regimes and complex tasks, establishing PAC-DP as a theoretically grounded framework for robot policy learning.

\end{abstract}


\section{Introduction}

Diffusion Policies (DPs) have emerged as a powerful framework for robotic imitation learning, leveraging iterative denoising to represent highly multimodal action distributions \cite{wolf2025diffusion,chi2025diffusion}. Despite their empirical success across complex manipulation tasks, the generalization properties of DPs remain relatively unexplored—particularly in low-data regimes where collecting expert demonstrations is costly and time-consuming. As a result, DPs are often trained on limited datasets, making generalization a central challenge.



Recent theoretical and empirical studies on Diffusion Models (DMs) suggest that in low-data training regimes minimizing an empirical denoising objective alone does not guarantee generalization in DMs. Li \emph{et al.} \cite{li2024good} showed that the optimal score function, obtained by minimizing the denoising objective, can lead DMs to generate samples closely tied to the training data. Moreover, DMs have been shown to memorize training samples when trained on limited datasets or with insufficient regularization \cite{bonnaire2026diffusion,gu2023memorization,baptista2025memorization}. Together, these findings suggest that minimizing the denoising objective alone may be insufficient to capture the underlying data distribution. We observed this problem also in DPs, which motivates us to derive learning principles with stronger generalization guarantees.

Existing approaches for improving the performance of DPs primarily rely on architectural inductive biases, structured conditioning mechanisms, or increasingly diverse demonstration datasets. Examples include geometry-aware scene representations \cite{ze20243d,ke20243d}, hierarchical and goal-conditioned diffusion policies \cite{xian2023chaineddiffuser,reuss2023goal}, and large-scale multi-task robot datasets \cite{team2024octo}. While these approaches have demonstrated empirical performance, they primarily improve generalization indirectly through architecture or data design, and provide limited theoretical guidance for designing the training objective itself.

\begin{figure*}[t]
    \centering
    \includegraphics[width=0.8\textwidth, trim={1.0cm 1.6cm 1.5cm 5cm},
    clip]{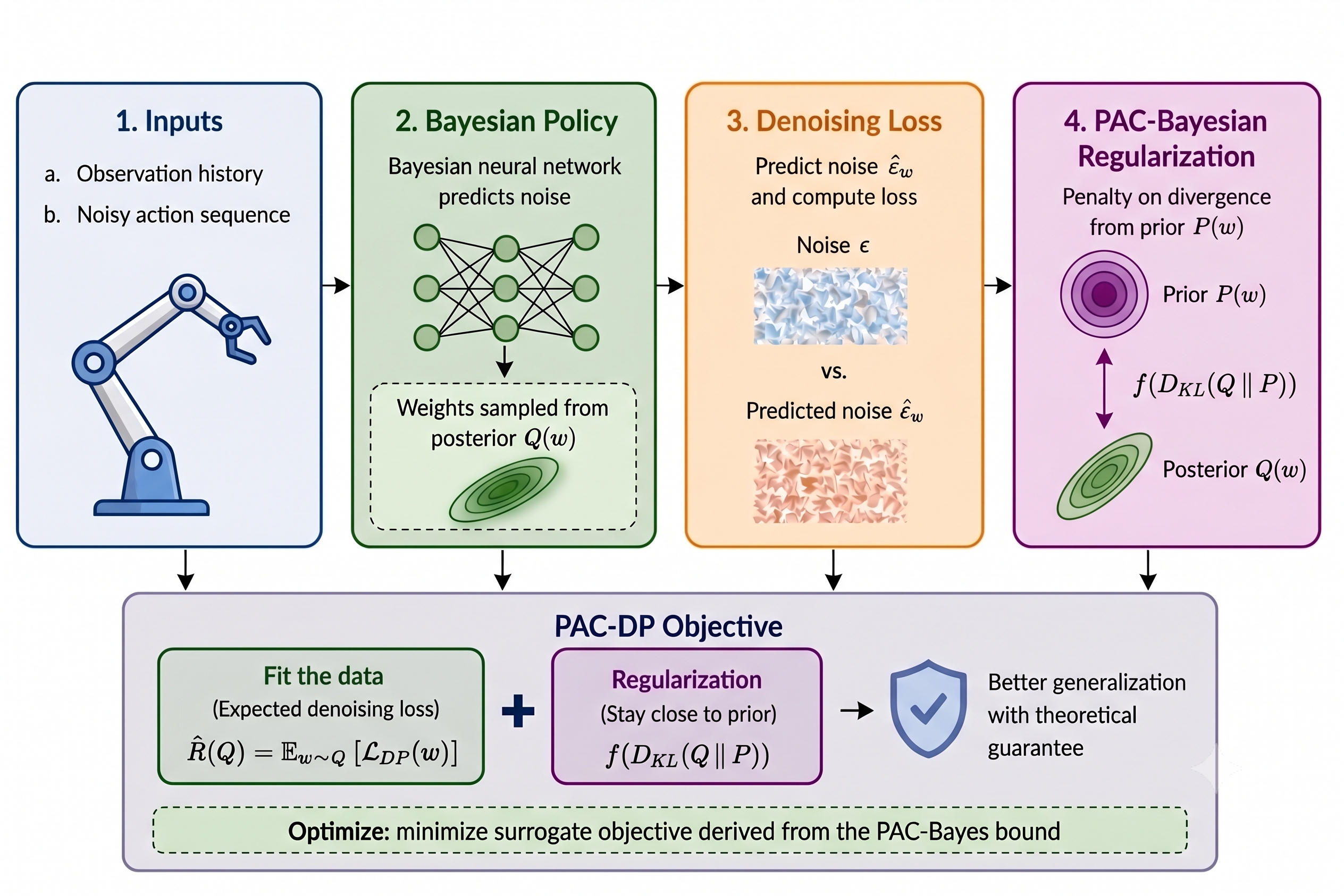}
    \vspace{-2mm}
    \caption{Overview of PAC-DP. The diffusion policy is modeled as a Bayesian neural network by learning a posterior distribution $Q(w)$ over network parameters. Training minimizes a PAC-Bayesian objective that combines the standard denoising loss with a KL-divergence regularizer between the posterior and prior distributions. During inference, actions are generated using either sampled network parameters or the posterior mean.}
    \label{fig:pac_diffusion_overview}
    \vspace{-3mm}
\end{figure*}

To address this challenge, we propose \textit{PAC-DP} (Fig.~\ref{fig:pac_diffusion_overview}), an approach that establishes theoretical guarantees on expected generalization performance for diffusion-based imitation learning. More in detail, we first derive a generalization bound on the training loss of DPs using Probably Approximately Correct-Bayesian (PAC-Bayesian) learning theory~\cite{alquier2024user}. The generalization bound consists of two terms, namely the standard empirical denoising loss~\cite{ho2020denoising} and a Kullback--Leibler (KL) divergence term. To transform the PAC-Bayes bound into a training objective, we propose to use a Bayesian neural network (BN)~\cite{perez2021tighter} to perform the denoising step in the DP. This allows us to minimize the generalization term depending on the KL divergence between the posterior and prior weight distributions. 
We evaluate the proposed framework from three complementary perspectives. First, denoising performance using the standard noise-prediction objective to assess score-matching quality; second, variational negative log-likelihood bounds to quantify generative modeling performance; and third, downstream robotic success rates on simulated manipulation benchmarks, focusing on low-data training regimes where regularization yields the greatest benefit. We finally validate real-world deployment of the proposed method on a physical Franka robot~\cite{haddadin2024franka}.

\noindent To summarize, the contributions of this paper are threefold:

\begin{itemize}

\item We formulate diffusion-based imitation learning within a PAC-Bayesian framework, deriving a novel training objective (PAC-DP) that mathematically couples the standard denoising loss with a KL-divergence regularizer.

\item We demonstrate that our objective yields superior denoising accuracy and a lower variational negative log-likelihood bound, indicating improved modeling of the action distribution.

\item We validate PAC-DP across the same manipulation benchmarks in \cite{chi2025diffusion} and a real robot, demonstrating higher robotic task success rates than standard diffusion policies, particularly in low-data regimes.

\end{itemize}
Together, these results establish PAC-Bayesian learning as a principled framework for improving the generalization of diffusion-based robot policies.

\section{Related Work}

\subsection{Diffusion-based Imitation Learning}

DP has become one of the dominant paradigms for robot imitation learning~\cite{chi2025diffusion}. Building on this foundation, subsequent work has primarily sought to improve policy generalization through architectural inductive biases and large-scale data.
A line of work improves spatial perception by modifying the policy architecture to incorporate explicit 3D scene representations. Geometry-aware diffusion policies replace purely image-based conditioning with representations such as sparse point clouds or tokenized 3D scene features, enabling more effective geometric reasoning and stronger generalization across viewpoints and spatial configurations~\cite{ke20243d,ze20243d}. Complementary approaches address long-horizon manipulation either through goal-conditioned diffusion policies that condition action generation on future goal observations~\cite{reuss2023goal} or through hierarchical architectures that decompose tasks into high-level macro-action prediction and low-level diffusion-based trajectory generation~\cite{xian2023chaineddiffuser}.
Alternatively, performance can be improved by scaling the amount and diversity of training data. Large scale robot foundation models pretrain vision-language-action policies on diverse datasets before adapting them to downstream tasks, substantially improving robustness and transferability across a wide range of manipulation tasks~\cite{team2024octo, kim2024openvla}.

Despite these advances, existing approaches have largely focused on improving policy performance through architectural innovations, richer representations, and larger training datasets. In contrast, comparatively little attention has been devoted to the design of theoretically grounded learning objectives for DPs. Our work complements these efforts by introducing a PAC-Bayesian regularization framework that improves generalization without modifying the underlying diffusion architecture.

\subsection{Generalization in Diffusion Models}

Generalization in DMs has attracted increasing attention from both generative modeling and privacy perspectives~\cite{zhang2025generalization,wen2024detecting}. Li et al.~\cite{li2024good} theoretically showed that, under finite-data training, minimizing the empirical denoising objective leads to empirical denoising error but does not prevent memorization, with the resulting model generating Gaussian-blurred replicas of training examples rather than genuinely novel images. 
Subsequent studies further investigated why such memorizing solutions are not always observed in practice. Bonnaire et al.~\cite{bonnaire2026diffusion} showed that DMs undergo a prolonged generalization phase before memorization emerges. They attributed this behavior to implicit dynamical regularization in the training dynamics. Similarly, Gu et al.~\cite{gu2023memorization} empirically showed that memorization becomes increasingly prominent under limited data, larger model capacity, and prolonged training. More recently, Baptista et al.~\cite{baptista2025memorization} developed a dynamical systems analysis showing that the empirical score-matching minimizer drives reverse probability-flow trajectories toward individual training samples. Based on this analysis, they argued that the principled regularization of the score estimation objective is necessary to prevent memorization.

While these studies have substantially improved our understanding of generalization in the diffusion-based image generation, there is little understanding of this phenomenon in DPs. Moreover, existing approaches have been developed specifically for the image generation setting, where they primarily characterize generalization behavior or leverage implicit regularization mechanisms rather than explicitly optimizing for generalization during training. We address this limitation by introducing a KL-divergence regularizer derived from a PAC-Bayes generalization bound, encouraging solutions that jointly minimize the empirical denoising loss while providing theoretical guarantees on generalization.

\subsection{PAC-Bayesian Learning in Robotics}

PAC-Bayesian learning provides a principled framework for analyzing the generalization of stochastic predictors by relating the expected test performance to empirical risk and the divergence between posterior and prior distributions over model parameters \cite{alquier2024user,perez2021tighter,biggs2023tighter}. Unlike classical uniform-convergence analyses, PAC-Bayesian bounds are data dependent and have been shown to yield meaningful generalization guarantees even for overparameterized neural networks \cite{perez2021tighter}.
Within robotics, PAC-Bayesian methods have been successfully applied to controller synthesis \cite{majumdar2018pac,majumdar2021pac}, motion planning under uncertainty \cite{veer2021probably}, imitation learning \cite{ren2021generalization}, and optimal control with stability guarantees \cite{boroujeni2024pac}. These works demonstrate the effectiveness of PAC-Bayesian analysis for reasoning about uncertainty and generalization in robotic systems.

Our work differs from prior research in two important aspects. First, we consider diffusion-based imitation learning, where policies are optimized through a denoising generative objective rather than conventional supervised losses. Second, rather than employing PAC-Bayes solely for post hoc generalization analysis, we derive a PAC-Bayesian regularized training objective and integrate it directly into diffusion optimization. To the best of our knowledge, this is the first work that leverages PAC-Bayesian theory as a regularization principle for training DPs.

\section{Method}

In this section, we first review the formulation of DP and then, building upon the standard DP loss, we show how PAC-Bayesian theory is used to derive PAC-DP. 

\subsection{Diffusion Policy Preliminaries}

\acs{dp}~\cite{chi2025diffusion} formulates imitation learning by modeling the conditional distribution of future robot action sequences using conditional Denoising Diffusion Probabilistic Models (DDPMs)~\cite{ho2020denoising}. Let ${x}_0 \in \mathbb{R}^{N \times T_a}$ denote a sequence of $T_a$ robot actions conditioned on the most recent $T_o$ observations $o \in \mathbb{R}^{M \times T_o}$. During training, the forward diffusion process progressively perturbs the action sequence with Gaussian noise $\mathcal{N}({0},{I})$ according to
\begin{equation}
{x}_t = \sqrt{\bar{\alpha}_t}{x}_0 + \sqrt{1-\bar{\alpha}_t}{\epsilon}, \qquad {\epsilon}\sim\mathcal{N}({0},{I}),
\label{eq:ddpm_forward}
\end{equation}
where $t\in\{1,\ldots,T\}$ denotes the diffusion timestep and $\bar{\alpha}_t$ is determined by a predefined noise schedule.

A neural network $\epsilon_\theta({x}_t,t,o)$ is trained to predict the injected noise by minimizing the denoising objective
\begin{equation}
\mathcal{L}_{\mathrm{DP}}(\theta) = \mathbb{E}_{{x}_0,t,{\epsilon}} \left[ \left\| {\epsilon} - \epsilon_\theta({x}_t,t,o) \right\|_2^2 \right],
\label{eq:ddpm_loss}
\end{equation}
where the expectation is taken over clean action sequences, diffusion timesteps, and Gaussian noise samples. Minimizing \eqref{eq:ddpm_loss}, we learn a denoising model $\epsilon_\theta$ that implicitly defines the reverse diffusion process for generating action sequences conditioned on the observations.
At inference time, actions are generated by iteratively denoising a Gaussian noise sample. Following~\cite{chi2025diffusion}, we employ Denoising Diffusion Implicit Models (DDIM)~\cite{song2020denoising} to accelerate sampling through deterministic reverse diffusion with substantially fewer denoising steps.

$\mathcal{L}_{\mathrm{DP}}$ in~\eqref{eq:ddpm_loss} is the standard training loss for diffusion policies. In the following subsection, we interpret this denoising objective as the empirical risk of a stochastic predictor, and show how it can be regularized within a PAC-Bayesian framework to obtain improved generalization guarantees.

\subsection{PAC-Bayesian Generalization Bounds}

PAC-Bayesian learning theory provides data-dependent upper bounds on the expected generalization error of stochastic predictors. Unlike conventional learning, which seeks for a single deterministic set of parameters, the PAC-Bayesian framework considers a distribution over model parameters. A predictor is obtained by sampling its parameters from this distribution, and the objective is to learn a posterior distribution whose expected performance generalizes well beyond the training data.

Let $\mathcal{S}=\{z_i\}_{i=1}^{N}$ denote a training dataset of $N$ i.i.d. samples drawn from an unknown distribution $\mathcal{D}$, and let $\ell(w,z)$ denote a bounded loss function. Given a prior distribution $P(w)$ over model parameters $w$, and a learned posterior distribution $Q(w)$, the empirical risk $\hat{R}(Q)$ and the population risk ${R}(Q)$ of the stochastic predictor are defined as
\begin{equation}
\hat{R}(Q) = \mathbb{E}_{w\sim Q} \left[ \frac{1}{N} \sum_{i=1}^{N} \ell(w,z_i) \right],
\end{equation}
and
\begin{equation}
R(Q) = \mathbb{E}_{w\sim Q} \left[ \mathbb{E}_{z\sim\mathcal D} \big[ \ell(w,z) \big]\right].
\end{equation}

To relate these two quantities, we employ the PAC-Bayes-kl inequality~\cite{seeger2002pac}. Using the relaxation obtained from Refined Pinsker's inequality~\cite{seldin2025machine}, the following bound holds with probability at least $1-\delta$:
\begin{equation}
\begin{aligned}
R(Q)\le & \hat R(Q) + \sqrt{ \frac{ 2\hat R(Q) \left( D_{\mathrm{KL}}(Q\|P) +\ln\frac{2\sqrt N}{\delta} \right)} {N} } \\ & + \frac{ 2 \left( D_{\mathrm{KL}}(Q\|P) + \ln\frac{2\sqrt N}{\delta} \right) }{N}.
\end{aligned}
\label{eq:pacbayes_bound}
\end{equation}
Equation~\eqref{eq:pacbayes_bound} highlights the central principle underlying PAC-Bayesian learning. The expected population risk is jointly controlled by the empirical risk and the divergence between the learned posterior and the prior distribution. Consequently, minimizing a surrogate objective derived from this bound naturally balances data fitting with the complexity of the learned posterior, providing a principled regularization mechanism for improving generalization.

In a general setting (the case of DP is detailed in Sec.~\ref{subsec:pac-dp}), we can instantiate this framework using a Bayesian neural network (BN)~\cite{blundell2015weight}. Hence, rather than optimizing a single deterministic weight vector, we learn a posterior distribution over the weights of the network. To retain tractability of the KL terms in~\eqref{eq:pacbayes_bound}, the prior $P$ and posterior $Q$ are parameterized as a factorized Gaussian
\begin{equation}
Q(w) = \mathcal N \left( \mu, \operatorname{diag}(\sigma^2) \right),
\label{eq:posterior}
\end{equation}
where the learnable variables are the posterior mean $\mu$ and the standard deviation $\sigma$. During training, network parameters are not optimized directly. Instead, the weights are sampled from the posterior using the reparameterization trick~\cite{kingma2014auto}, i.e.,
\begin{equation}
w = \mu + \sigma \odot {\xi}, \qquad {\xi} \sim \mathcal N({0},{I}),
\label{eq:reparam}
\end{equation}
where $\odot$ denotes element-wise multiplication. This enables the optimization of $(\mu,\sigma)$ via backpropagation. To ensure non-negative standard deviations, we parameterize~\cite{perez2021tighter}
\begin{equation}
\sigma = \log(1+\exp(\rho)),
\label{eq:sigma}
\end{equation}
where $\rho$ is an unconstrained learnable parameter.

\subsection{PAC-DP}\label{subsec:pac-dp}

To instantiate PAC-DP, we model the noise prediction network used in DP as the BN introduced in the previous subsection. Consequently, the deterministic network weights $\theta$ in~\eqref{eq:ddpm_loss} are replaced by the stochastic weights $w$ sampled from the learned posterior distribution $Q(w)$. This leads to the Bayesian noise predicting network $\epsilon_w(x_t,t,o)$.
The empirical risk of the stochastic predictor is then given by the expectation of the diffusion objective over the posterior, i.e.,
\begin{equation}
\hat{R}(Q) = \mathbb{E}_{w\sim Q} \left[ \mathcal{L}_{\mathrm{DP}}(w) \right].
\label{eq:empirical_risk_dp}
\end{equation}
Equation~\eqref{eq:empirical_risk_dp} establishes the key connection between DP training and PAC-Bayesian theory by interpreting the standard denoising objective as the empirical risk of a stochastic predictor.

Substituting \eqref{eq:empirical_risk_dp} into the PAC-Bayes bound \eqref{eq:pacbayes_bound} yields a learning objective consisting of an empirical denoising term and a KL-divergence regularizer. Following common practice in PAC-Bayesian optimization, we introduce a weighting coefficient $\lambda$ to control the strength of the regularization term and minimize the following differentiable surrogate objective:
\begin{equation}
\begin{aligned}
\mathcal{L}_{\mathrm{PAC-DP}} = & \hat R(Q) + \sqrt{ \frac{ 2\hat R(Q) \left( \lambda D_{\mathrm{KL}}(Q\|P) + \ln\frac{2\sqrt N}{\delta} \right)} {N} } \\[2mm] & + \frac{ 2 \left( \lambda D_{\mathrm{KL}}(Q\|P) + \ln\frac{2\sqrt N}{\delta} \right) }{N}.
\end{aligned}
\label{eq:pacd_loss}
\end{equation}
The proposed objective balances accurate denoising ($\hat{R}(Q)$) with a regularization term derived directly from the PAC-Bayes generalization bound. The KL term has a clear learning-theoretic interpretation, encouraging posterior distributions that achieve low empirical risk while remaining close to the prior.

The PAC-Bayes-kl inequality formally assumes bounded loss functions, whereas $\mathcal{L}_{\mathrm{DP}}$ in~\eqref{eq:ddpm_loss} is theoretically unbounded because it is defined by a squared prediction error. A common theoretical remedy is to consider the clipped loss $\ell_{\mathrm{clip}}=\min(\ell,1)$, which satisfies the assumptions required by the bound. In practice, however, the denoising loss rapidly decreases below one during the first few training iterations, and remains in this regime throughout optimization. Consequently, the clipped and original objectives become nearly identical after the initial training stage. Motivated by this observation, we optimize the original denoising objective while retaining the PAC-Bayesian regularization term.

Because both the prior and posterior are factorized Gaussian distributions, the KL divergence admits a closed-form expression that can be computed independently for each Bayesian layer, i.e.,
\begin{equation}
D_{\mathrm{KL}}(Q\|P) = \sum_{l=1}^{L} D_{\mathrm{KL}}(Q_l\|P_l),
\label{eq:layerwise_kl}
\end{equation}
where $L$ denotes the number of Bayesian layers.
The posterior parameters $(\mu,\sigma)$ are optimized using stochastic gradient descent. For each forward pass, network parameters are sampled from the posterior using \eqref{eq:reparam}, the PAC-DP objective \eqref{eq:pacd_loss} is evaluated, and gradients are back-propagated to update the posterior parameters. During inference, the learned posterior may either be replaced by its mean to obtain a deterministic policy or sampled to preserve stochasticity. The complete training and inference procedures are summarized in Algorithms~\ref{alg:training} and~\ref{alg:inference}, respectively.

\begin{algorithm}[t]
\caption{PAC-DP Training}
\label{alg:training}
\begin{algorithmic}[1]
\REQUIRE Dataset $\mathcal{S}$, prior $P$, posterior $Q$
\WHILE{not converged}
    \STATE Sample $({x}_0, o)$, timestep $t$, and noise ${\epsilon}$
    \STATE Construct noisy action sequence ${x}_t$ using \eqref{eq:ddpm_forward}
    \STATE Sample policy parameters $w \sim Q$ using \eqref{eq:reparam}
    \STATE Predict $\hat{{\epsilon}} \leftarrow \epsilon_w({x}_t, t, o)$
    \STATE Compute loss $\mathcal{L} \leftarrow \mathcal{L}_{\mathrm{PAC-DP}}$ via \eqref{eq:pacd_loss}
    \STATE Update $Q$ parameters $(\mu, \rho)$ via $\nabla \mathcal{L}$
\ENDWHILE
\end{algorithmic}
\end{algorithm}

\begin{algorithm}[t]
\caption{PAC-DP Inference}
\label{alg:inference}
\begin{algorithmic}[1]
\REQUIRE Observations $o$, posterior $Q$, inference mode $m$
\IF{$m = \texttt{mean}$}
    \STATE $w \leftarrow \mathbb{E}[Q] = \mu$
\ELSE
    \STATE Sample $w \sim Q$
\ENDIF
\STATE Initialize ${x}_T \sim \mathcal{N}({0},{I})$
\FOR{$t = T, \ldots, 1$}
    \STATE Predict noise $\hat{{\epsilon}} \leftarrow \epsilon_w({x}_t, c, t)$
    \STATE Compute ${x}_{t-1}$ using the DDIM scheduler
\ENDFOR
\STATE \textbf{return} ${x}_0$
\end{algorithmic}
\end{algorithm}

\section{Evaluation}

\label{sec:setup}
We evaluate PAC-DP against two baselines:
\begin{itemize}
    \item \textit{\acs{dp}} \cite{chi2025diffusion}: The standard DP utilizing a deterministic network trained via the standard denoising loss.
    \item \textit{Bayesian DP (BDP)}: A BN is used for noise prediction, but it is optimized using the standard denoising objective~\eqref{eq:empirical_risk_dp} without PAC-Bayesian regularization. This ablation of PAC-DP separates the effect of stochastic parameterization from regularization. 
\end{itemize}
For inference, we deploy \textit{mean inference}, which uses the deterministic posterior mean $\mu$.

\subsection{Simulation Environments and Setup}
\begin{figure*}[t]
    \centering
    \includegraphics[trim={1.0cm 4cm 0.5cm 2.4cm}, clip, width=\textwidth]{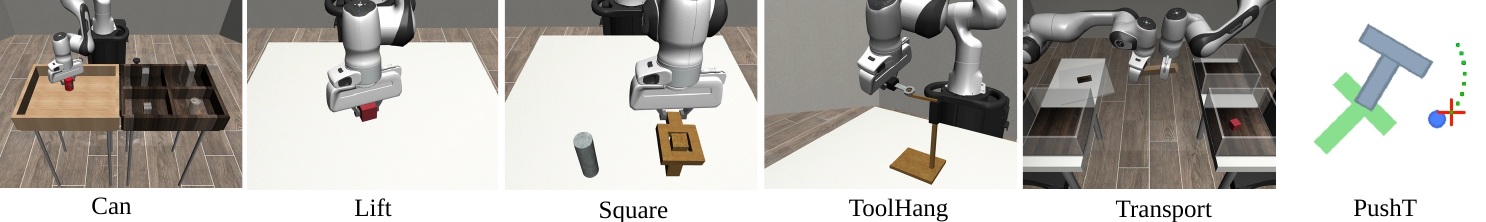}
    \caption{Adopted benchmarks from~\cite{chi2025diffusion} used for simulation experiments.}
    \label{fig:benchmarks}
\end{figure*}

Evaluations are conducted on the \acs{dp} benchmark suite~\cite{chi2025diffusion} (Fig.~\ref{fig:benchmarks}), containing the \texttt{PushT} task \cite{florence2022implicit} and five Robomimic tasks (\texttt{Lift}, \texttt{Can}, \texttt{Square}, \texttt{Transport}, \texttt{ToolHang}) \cite{mandlekar2021matters}. Experiments utilize state-based observations, successful expert demonstrations, and DDIM sampling with 10 denoising steps.

We adopt the posterior parameterization and factorized Gaussian prior configuration of \cite{biggs2023tighter}. Prior mean parameters are initialized from a truncated zero-mean Gaussian ($\sigma = 1/\sqrt{n_{\mathrm{in}}}$, where $n_{\mathrm{in}}$ is the dimension of the inputs to a particular layer, truncated at $\pm2$ standard deviations). For the posterior, the scale parameter $\rho$ in~\eqref{eq:sigma} is chosen as a constant hyperparameter. We initialize $Q = P$ during the training. For noise prediction, we use the same U-Net architecture as in~\cite{chi2025diffusion}, but converting all layers to variational Bayesian variants except for the conditional projection layers (see Sec.~\ref{sec:ablation-network} for the ablation study).

We consider the following evaluation metrics:
\begin{itemize}
    \item \textit{Noise Prediction Error (NPE)}: Measured on a held-out test dataset to quantify validation overfitting.
    \item \textit{Negative Log-Likelihood (NLL) Lower Bound}: Evaluated on the held-out test dataset using the information-theoretic formulation of \cite{kong2023information} to assess distribution-level generalization. Since we use a discrete-time diffusion model with a cosine variance schedule, computing the exact continuous log-likelihood is intractable.
    \item \textit{Success Rate (SR)}: Computed across $1,000$ randomly initialized environments ($30,000$ total rollouts per task across all seeds) to provide a stable estimate of converged task-level generalization.
\end{itemize}
Unless otherwise stated, all reported metrics are averaged over the last 10 checkpoints (saved every $5,000$ steps) across 3 random seeds.

\subsection{Results}

\subsubsection{{\texttt{PushT}}}
\label{sec:results-pusht}

\begin{figure*}[t]
    \centering
    \includegraphics[width=\textwidth]{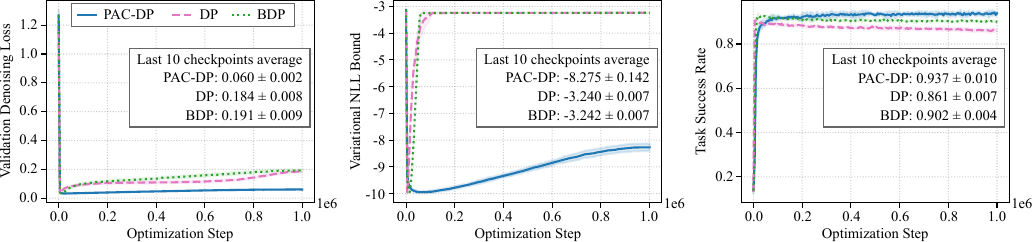}
    \caption{Optimization dynamics on the \texttt{PushT} task. Shaded regions indicate the std over three random seeds. Reported values correspond to the mean $\pm$ std over the final ten checkpoints.}
    \label{fig:pusht_metrics}
\end{figure*}

We separately analyze \texttt{PushT} to report a set of results showing that, as the training evolves, \acs{dp} and \acs{bdp} tend to overfit the training data.
Fig.~\ref{fig:pusht_metrics} compares the optimization dynamics of PAC-DP, \acs{dp}, and \acs{bdp} using 200 demonstrations, 140 (70\%) for training and 60 (30\%) for the validation of the NPE and NLL lower bound. All models are trained for $10^6$ optimization steps and evaluated every 5,000 iterations. PAC-DP hyperparameters are fixed to $\rho=-4.5$ with $\lambda=10^{-4}$.

The left panel reports NPE. Although all methods rapidly reduce the training objective during the early stages of optimization, \acs{dp} and \acs{bdp} exhibit a gradual increase in NPE as training continues, indicating degradation in generalization. In contrast, PAC-DP maintains a consistently low NPE throughout optimization, converging to $0.060\pm0.002$, compared with $0.184\pm0.008$ for \acs{dp} and $0.191\pm0.009$ for \acs{bdp}. This corresponds to approximately a 67\% reduction in NPE relative to the \acs{dp}.

The middle panel reports the variational NLL bound~\cite{kong2023information}, which evaluates the quality of the learned diffusion model. Lower values correspond to tighter likelihood bounds and indicate a more informative denoiser. While \acs{dp} and \acs{bdp} initially improve the bound, both methods quickly saturate near $-3.24$, which corresponds to a random Gaussian-denoiser~\cite{kong2023information}. In contrast, PAC-DP continues to maintain a substantially tighter bound throughout training, converging to $-8.275\pm0.142$. These results suggest that the PAC-Bayesian regularization objective guides optimization toward solutions that retain more information about the underlying demonstration distribution. This observation is also consistent with recent results on DMs~\cite{li2024good}, showing that optimizing the denoising objective alone can admit empirical solutions that remain closely tied to the training data under limited-data settings.

These improvements translate to downstream control performance. As shown in the right panel, PAC-DP converges to a stable task SR of $93.7\%\pm1.0\%$, while both baselines exhibit a gradual decline in success rate after the early stages of training (\acs{dp} ends at $86.1\%\pm0.7\%$, \acs{bdp} at $90.2\%\pm0.4\%$). This indicates that improved generalization of the denoising model can translates into improved policy execution.

It is worth highlight that \acs{bdp} (an ablation of PAC-DP) improves over \acs{dp} in terms of success rate, but not in the other metrics, while PAC-DP outperforms both baselines. This indicates that the combination of Bayesian parameterization and PAC-Bayesian regularization proposed in PAC-DP is needed to maximize performance. We further investigate this point as follows. 

\subsubsection{{Robomimic Benchmark}}

\begin{table*}[t]
\centering
\caption{Comparison of validation performance across Robomimic benchmark. 
}
\label{tab:generalization_metrics_all_tasks}
\resizebox{\textwidth}{!}{
\begin{tabular}{lccccccccc}
\toprule

& \multicolumn{3}{c}{Noise Prediction Error (NPE) $\downarrow$}
& \multicolumn{3}{c}{Variational NLL $\downarrow$}
& \multicolumn{3}{c}{Success Rate (SR) $\uparrow$}\\

\cmidrule(lr){2-4}
\cmidrule(lr){5-7}
\cmidrule(lr){8-10}

Task & DP & BDP & PAC-DP & DP & BDP & PAC-DP & DP & BDP & PAC-DP
\\
\midrule
\texttt{Lift} & 0.175 $\pm$ 0.008 & 0.177 $\pm$ 0.002 & \textbf{0.036 $\pm$ 0.001}& -5.182 $\pm$ 0.033 & -5.201 $\pm$ 0.022 & \textbf{-13.012 $\pm$ 0.512} & 0.999 $\pm$ 0.002 & 0.999 $\pm$ 0.002 & \textbf{0.999 $\pm$ 0.001} \\

\texttt{Can} & 0.106 $\pm$ 0.003 & 0.074 $\pm$ 0.006 & \textbf{0.021 $\pm$ 0.001}& -4.712 $\pm$ 0.025 & -4.754 $\pm$ 0.015 & \textbf{-13.521 $\pm$ 0.741} & 0.990 $\pm$ 0.004 & 0.988 $\pm$ 0.004 & \textbf{0.990 $\pm$ 0.001} \\

\texttt{Square} & 0.093 $\pm$ 0.001& 0.063 $\pm$ 0.001 & \textbf{0.038 $\pm$ 0.003} & -4.824 $\pm$ 0.025 & -4.917 $\pm$ 0.027& \textbf{-8.619 $\pm$ 0.310} & \textbf{0.920 $\pm$ 0.005}& 0.917 $\pm$ 0.003& 0.900 $\pm$ 0.003\\

\texttt{Transport} & 0.062 $\pm$ 0.002 & 0.048 $\pm$ 0.002 & \textbf{0.022 $\pm$ 0.001}& -5.615 $\pm$ 0.014 & -5.734 $\pm$ 0.013 &\textbf{-11.587 $\pm$ 0.624}& 0.761 $\pm$ 0.035 & \textbf{0.768 $\pm$ 0.021}& 0.701 $\pm$ 0.025\\

\texttt{ToolHang} & 0.065 $\pm$ 0.003& 0.048 $\pm$ 0.001& \textbf{0.013 $\pm$ 0.001} & -5.014 $\pm$ 0.032 & -5.106 $\pm$ 0.025 & \textbf{-16.383 $\pm$ 0.117}& 0.401 $\pm$ 0.046& 0.544 $\pm$ 0.053& \textbf{0.738 $\pm$ 0.011} \\

\bottomrule
\end{tabular}
}
\end{table*}

Table~\ref{tab:generalization_metrics_all_tasks} summarizes obtained results across different tasks using 200 demonstrations, with 30\% of the data held out for the validation of the NPE and NLL lower bound (i.e., the models are trained on 140 demonstrations). Tasks are trained for $2.5\times10^5$ gradient steps, except \texttt{Transport} and \texttt{ToolHang}, which require $5\times10^5$ steps for convergence. Hyperparameters are fixed to $\rho=-4.5$, $\lambda=10^{-5}$ for \texttt{Lift}, \texttt{Can}, and \texttt{ToolHang}, and $\lambda=10^{-6}$ for \texttt{Square} and \texttt{Transport} (see the sensitivity analysis at the end of this section).

PAC-DP consistently achieves the lowest NPE across all tasks, outperforming both \acs{dp} and \acs{bdp}. Similar trends are observed for the variational NLL lower bound, indicating improved probabilistic modeling and better generalization of the learned denoising process.
The improvements in denoising performance also translate to improved downstream performance for several tasks. In particular, PAC-DP achieves the largest gain on the challenging \texttt{ToolHang} task, increasing the success rate from $0.406$ (\acs{dp}) and $0.544$ (\acs{bdp}) to $0.738$. On the simpler \texttt{Lift} and \texttt{Can} tasks, all methods already achieve near-perfect performance, leaving little room for further improvement despite the significant reduction in denoising error.

For the \texttt{Square} and \texttt{Transport} tasks, however, reduced NPE does not directly translate into higher SR. This is expected since PAC-DP optimizes a PAC-Bayes bound on the diffusion denoising objective rather than on the task SR itself. Consequently, improvements in the learned noise predictor do not necessarily yield proportional gains in downstream control performance.

\subsubsection{{Low-data Regime Evaluation}}
\label{sec:results-low-regime}

The objective of this experiment is to evaluate policy learning in finite-data regimes, where regularization is expected to provide the greatest benefit. Models are trained using $20$, $40$, $80$, and $200$ demonstrations for all tasks except \texttt{Transport} and \texttt{ToolHang}, which are evaluated only with $200$ demonstrations since the baseline policies fail to learn these tasks reliably with fewer demonstrations. PAC-DP results are obtained using $\lambda \in \{10^{5}, 10^{6}\}$ and $\rho \in \{ -3.5, -4.0, -4.5, -5.0\}$ (see the hyperparameter sensitivity analysis in the next subsection). 

Table~\ref{tab:success_rate_comparison} shows the obtained results. With 20 demonstrations, PAC-DP outperforms the baselines in all tasks. The highest improvement is observed for \texttt{PushT} and \texttt{Can}. As the number of demonstrations increases, the performance gap generally decreases, suggesting that the proposed regularization is particularly effective when data are limited. PAC-DP outperforms the baselines in the majority of the tests. An exception is the \texttt{Square} task, where DP achieves slightly higher success rates for 40 and 80 demonstrations. As mentioned in Sec.~\ref{sec:results-pusht}, this result is expected in simpler tasks where reduced NPE does not necessarily lead to improved task SR. However, the performance differences are relatively small and disappear at 200 demonstrations. It is worth mentioning that PAC-DP significantly outperforms the baselines for the \texttt{ToolHang} also with 200 demonstrations (85.8\% SR against 64.3\% for \acs{dp} and 72.0\% for \acs{bdp}). This is probably due to the complexity of this task requiring a long sequence of fine manipulations, and for which a principled regularization becomes increasingly beneficial. 


Overall, we observe that PAC-DP outperforms or matches the baseline on the evaluated tasks while providing the largest improvements in finite-data regimes and on more challenging manipulation tasks. Combined with the improved denoising and likelihood metrics presented in Fig.~\ref{fig:pusht_metrics}, these results support the hypothesis that PAC-Bayesian regularization improves the generalization of diffusion policies, and that the reduced NPE can translate into higher task success.

\begin{table*}[t]
\caption{Task SR (mean (std)) under different numbers of training demonstrations. 
}
\label{tab:success_rate_comparison}
\centering
\setlength{\tabcolsep}{3pt}
\renewcommand{\arraystretch}{1.15}
\resizebox{\textwidth}{!}{%
\begin{tabular}{lcccccccccccc}
\toprule
&
\multicolumn{3}{c}{20 Demonstrations} &
\multicolumn{3}{c}{40 Demonstrations} &
\multicolumn{3}{c}{80 Demonstrations} &
\multicolumn{3}{c}{200 Demonstrations} \\

\cmidrule(lr){2-4}
\cmidrule(lr){5-7}
\cmidrule(lr){8-10}
\cmidrule(lr){11-13}

{Task}
& {DP} & {BDP} & {PAC-DP}
& {DP} & {BDP} & {PAC-DP}
& {DP} & {BDP} & {PAC-DP}
& {DP} & {BDP} & {PAC-DP} \\

\midrule

\texttt{Push-T}
& 0.481 (0.044) & 0.495 (0.037)& \textbf{0.549 (0.021)}
& 0.532 (0.027) & 0.552 (0.074)& \textbf{0.608 (0.014)}
& 0.705 (0.024) & 0.734 (0.029) & \textbf{0.805 (0.025)}
& 0.901 (0.033) & 0.926 (0.017) & \textbf{0.956 (0.016)}\\

\texttt{Lift}
& 0.972 (0.002) & 0.990 (0.005) & \textbf{0.996 (0.004)}
& 0.991 (0.002) & 0.994 (0.006) & \textbf{0.997 (0.004)}
& 0.998 (0.002) & 0.998 (0.002) & 0.998 (0.002)
& \textbf{0.999 (0.001)} & 0.999 (0.002) & \textbf{0.999 (0.001)}\\

\texttt{Can}
& 0.781 (0.017) & 0.865 (0.036) & \textbf{0.906 (0.022)}
& 0.923 (0.044) & 0.925 (0.023) & \textbf{0.964 (0.041)}
& 0.984 (0.003) & 0.986 (0.007)& \textbf{0.990 (0.005)}
& 0.992 (0.005) & 0.993 (0.002)& \textbf{0.995 (0.003)}\\

\texttt{Square}
& 0.470 (0.026)& 0.474 (0.032)& \textbf{0.500 (0.011)}
& \textbf{0.701 (0.064)}& 0.700 (0.021) & 0.686 (0.023)
& \textbf{0.882 (0.041)}& 0.851 (0.043) & 0.877 (0.019)
& 0.931 (0.003)& 0.923 (0.006) & \textbf{0.945 (0.004)} \\

\texttt{Transport}
& -- & -- & 
& -- & -- &
& -- & -- &
& 0.795 (0.054) & 0.786 (0.026) & \textbf{0.806 (0.019)} \\

\texttt{ToolHang}
& -- & -- &
& -- & -- &
& -- & -- &
& 0.643 (0.041) & 0.720 (0.022) & \textbf{0.858 (0.025)} \\

\bottomrule
\end{tabular}
}
\end{table*}

\subsubsection{{Hyperparameter Sensitivity Analysis}}
We evaluate the sensitivity of PAC-DP to the two hyperparameters $\lambda$ and $\rho$. Table~\ref{tab:hyperparameter_ablation} reports the average success rate over all hyperparameter configurations together with the best-performing configuration. For comparison, we also report the performance of \acs{bdp}, obtained from PAC-DP with $\lambda=0$.
During preliminary experiments, we observed that $\lambda > 10^{-5}$ consistently degraded performance in Robomimic benchnark. In this regime, the KL-divergence term dominates the loss~\eqref{eq:pacd_loss}, forcing the posterior to remain overly close to the prior.
Consequently, the ablation considers only $\lambda\in\{10^{-5},10^{-6}\}$.
For the posterior initialization, we evaluated $\rho\in\{-3.5,-4.0,-4.5,-5.0\}$. Larger values of $\rho$ increase the variance of sampled network parameters, leading to noisy gradient estimates and unstable training. 

Overall, PAC-DP performs consistently across a broad range of hyperparameter configurations. For most tasks, the average performance is close to that of the best-performing configuration, indicating that the proposed method does not rely on highly specific parameter tuning. The largest performance gaps are observed on the more challenging tasks, particularly \texttt{Transport}, where selecting an appropriate regularization weight provides additional gains.
The comparison with \acs{bdp} further highlights the importance of the PAC-Bayesian regularizer. Although both methods employ the same network architecture, PAC-DP consistently achieves higher average SR across almost all tasks and demonstration settings. The improvement is particularly pronounced on the more difficult manipulation tasks. For example, on \texttt{Transport}, PAC-DP increases the average success rate from $55.0$\% to $67.4$\% while substantially reducing the variability across hyperparameter configurations ($28.0$\% vs. $9.2$\% std). Similarly, on \texttt{ToolHang}, the average success rate improves from $67.9$\% to $82.6$\%. These results indicate that the performance gains arise primarily from the PAC-Bayesian regularizer rather than from the BN alone.


\begin{table}[t]
\centering
\caption{Hyperparameter sensitivity of PAC-DP. Results are averaged over all PAC-DP configurations. For comparison, BDP reports the average over the same $\rho$ values.}
\label{tab:hyperparameter_ablation}
\resizebox{\columnwidth}{!}{
\begin{tabular}{llccc}
\toprule
{Task} &
{\# Demos} &
{BDP Avg.} &
{PAC-DP. Avg.} &
{PAC-DP. Best} \\
\midrule

\multirow{4}{*}{\texttt{PushT}}
&20  & 0.471 $\pm$ 0.022 & 0.507 $\pm$ 0.030 & \textbf{0.549 $\pm$ 0.021} \\
&40  & 0.537 $\pm$ 0.013 & 0.577 $\pm$ 0.020 & \textbf{0.608 $\pm$ 0.014} \\
&80  & 0.719 $\pm$ 0.019 & 0.741 $\pm$ 0.039 & \textbf{0.805 $\pm$ 0.025} \\
&200 & 0.917 $\pm$ 0.011 & 0.941 $\pm$ 0.003 & \textbf{0.956 $\pm$ 0.016} \\

\midrule

\multirow{4}{*}{\texttt{Lift}}
&20  & 0.985 $\pm$ 0.004 & 0.990 $\pm$ 0.005 & \textbf{0.996 $\pm$ 0.004} \\
&40  & 0.989 $\pm$ 0.006 & 0.992 $\pm$ 0.008 & \textbf{0.997 $\pm$ 0.004} \\
&80  & 0.996 $\pm$ 0.002 & 0.997 $\pm$ 0.001 & \textbf{0.998 $\pm$ 0.002} \\
&200 & 0.998 $\pm$ 0.001 & 0.998 $\pm$ 0.000 & \textbf{0.999 $\pm$ 0.001} \\

\midrule

\multirow{4}{*}{\texttt{Can}}
&20  & 0.859 $\pm$ 0.035 & 0.868 $\pm$ 0.020 & \textbf{0.906 $\pm$ 0.022} \\
&40  & 0.914 $\pm$ 0.011 & 0.931 $\pm$ 0.018 & \textbf{0.964 $\pm$ 0.041} \\
&80  & 0.978 $\pm$ 0.004 & 0.978 $\pm$ 0.006 & \textbf{0.990 $\pm$ 0.005} \\
&200 & 0.992 $\pm$ 0.001 & 0.991 $\pm$ 0.004 & \textbf{0.995 $\pm$ 0.003} \\

\midrule

\multirow{4}{*}{\texttt{Square}}
&20  & 0.449 $\pm$ 0.022 & 0.452 $\pm$ 0.030 & \textbf{0.500 $\pm$ 0.011} \\
&40  & 0.664 $\pm$ 0.029 & 0.666 $\pm$ 0.010 & \textbf{0.686 $\pm$ 0.023} \\
&80  & 0.808 $\pm$ 0.060 & 0.816 $\pm$ 0.033 & \textbf{0.877 $\pm$ 0.019} \\
&200 & 0.883 $\pm$ 0.059 & 0.919 $\pm$ 0.024 & \textbf{0.945 $\pm$ 0.004} \\

\midrule

\texttt{Transport}
&200 & 0.550 $\pm$ 0.280 & 0.674 $\pm$ 0.092 & \textbf{0.806 $\pm$ 0.019} \\

\texttt{ToolHang}
&200 & 0.679 $\pm$ 0.043 & 0.826 $\pm$ 0.024 & \textbf{0.858 $\pm$ 0.025} \\

\bottomrule
\end{tabular}
}
\end{table}

\subsubsection{{BN Ablations}}
\label{sec:ablation-network}

Table~\ref{tab:conditioning_ablation} compares two Bayesian network architectures on the \texttt{ToolHang} task. The first treats both the denoising and conditioning networks as Bayesian (Full BN), while the proposed architecture (Ours) models only the denoising network as Bayesian and keeps the conditioning pathway deterministic.

Across all combinations of $\lambda$ and $\rho$, the proposed architecture consistently outperforms the Full BN model, improving the success rate from at most $27.8$\% for Full BN to up to $85.8$\% for ours. This trend remains consistent for both values of $\lambda$, indicating that the performance difference is primarily due to the network architecture rather than the regularization weight.
These results suggest that introducing parameter uncertainty into the conditioning layers degrades the quality of the observation representation used during denoising. In contrast, keeping the conditioning network deterministic provides a stable conditioning signal while allowing the denoising network to benefit from Bayesian regularization. 

\begin{table}[t]
\centering
\caption{SR (mean $\pm$ std) on the \texttt{ToolHang} task for different hyperparameters and different BN architectures.}
\label{tab:conditioning_ablation}
\resizebox{\columnwidth}{!}{
\begin{tabular}{ccccc}
\toprule

&
\multicolumn{2}{c}{$\lambda=10^{-5}$}
&
\multicolumn{2}{c}{$\lambda=10^{-6}$}
\\

$\rho$
& Full BN & Ours & Full BN & Ours
\\
\midrule
-3.5 & 0.016 $\pm$ 0.005 & \textbf{0.815 $\pm$ 0.041} & 0.144 $\pm$ 0.028 & \textbf{0.813 $\pm$ 0.034}\\
-4.0 & 0.115 $\pm$ 0.028 & \textbf{0.838 $\pm$ 0.026} & 0.257 $\pm$ 0.007 & \textbf{0.827 $\pm$ 0.023}\\
-4.5 & 0.166 $\pm$ 0.017 & \textbf{0.858 $\pm$ 0.019} & 0.251 $\pm$ 0.021 & \textbf{0.830 $\pm$ 0.026}\\
-5.0 & 0.179 $\pm$ 0.031 & \textbf{0.844 $\pm$ 0.026} & 0.278 $\pm$ 0.023 & \textbf{0.779 $\pm$ 0.021}\\
\bottomrule
\end{tabular}
}
\end{table}

\subsection{Real-World Experiments}

We further evaluate PAC-DP on a real Franka Emika Panda robot to verify that the learned policies can be successfully deployed on physical hardware. Four representative manipulation tasks are considered: \texttt{Lift}, \texttt{Can}, \texttt{Square}, and \texttt{ToolHang}.
For each task, we generate five trajectories from different initial object configurations using the trained policy. Each trajectory is then executed ten times, resulting in a total of fifty executions per task. The learned policies are executed in an open-loop manner without replanning or online feedback correction.

\begin{figure}[t]
    \centering
    \begin{overpic}[trim={0cm 1.7cm 0cm 0.5cm},clip,width=\columnwidth]{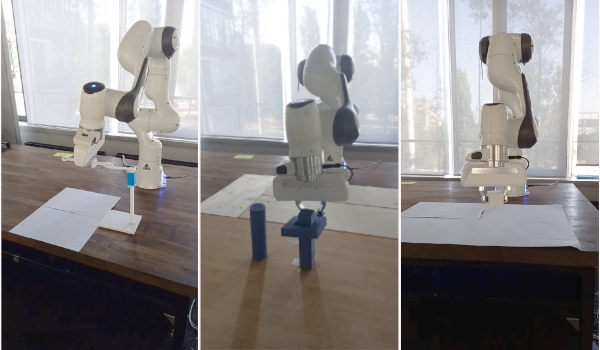}
        \put(7,2){\textcolor{white}{\texttt{ToolHang}}}
        \put(43,2){\textcolor{white}{\texttt{Square}}}
        \put(78,2){\textcolor{white}{\texttt{Lift}}}
    \end{overpic}
    \caption{Representative executions of PAC-DPs on a real robot.}
    \label{fig:real_world}
\end{figure}

PAC-DP successfully completes all executions across all four tasks, achieving a $100\%$ success rate over the 200 real-world trials. Representative execution sequences are shown in Fig.~\ref{fig:real_world}. These results demonstrate that the PAC-DP policies can be transferred directly from simulation to a physical robot without sacrificing execution reliability. Although these experiments are not intended as a comprehensive real-world benchmark, they confirm that the proposed regularization framework produces practically deployable diffusion policies capable of robust manipulation on real hardware.

\section{Conclusion and Future Work}

This paper introduced \emph{PAC-DP}, a PAC-Bayesian framework for diffusion-based imitation learning. By modeling the diffusion policy as a Bayesian neural network, we derived a training objective that augments the standard denoising loss with a PAC-Bayesian complexity regularizer, establishing a principled connection between diffusion policy optimization and statistical learning theory. The proposed learning objective is directly derived from a PAC-Bayes generalization bound, encouraging policies that jointly minimize empirical denoising error and a PAC-Bayesian complexity term.
Experimental evaluations on multiple robotic manipulation benchmarks demonstrated that PAC-DP consistently improves denoising performance and variational negative log-likelihood while yielding higher task success rates, particularly on challenging tasks and in low-data training regimes. Ablation studies further showed that these gains stem from the proposed PAC-Bayesian regularization objective rather than Bayesian parameterization alone, and real-world experiments confirmed successful deployment on a physical Franka robot.

Future work will investigate data-dependent PAC-Bayesian priors to derive tighter and less conservative generalization bounds. An additional direction is to extend the proposed framework to vision-based and multi-task diffusion policies, where stronger regularization may provide even greater benefits in large-scale robotic learning.




\bibliography{Reference}
\bibliographystyle{IEEEtran}

\end{document}